\documentclass[10pt,twocolumn,letterpaper]{article}

\usepackage[pagenumbers]{cvpr} 

\usepackage{cuted} 

\usepackage{currfile} 

\usepackage{caption} 
\definecolor{cvprblue}{rgb}{0.21,0.49,0.74}
\usepackage[pagebackref,breaklinks,colorlinks,allcolors=cvprblue]{hyperref}

\def\paperID{*****} 
\def\confName{CVPR}
\def\confYear{2026}

\title{Fursee: Hybrid YOLO-DINOv3 Framework for Fursuit Identity Retrieval and Clustering}

\author{Jundi Wu\\
Shandong University\\
Qingdao, China\\
{\tt\small lionking@wjd.email}
}

\begin{document}
\maketitle
\begin{abstract}
Global furry conventions produce massive fursuit photographs, while manual sorting brings heavy labor costs and calls for automatic identity retrieval and clustering solutions. General multimodal models lack dedicated optimization for complex fursuit scenes, and no public benchmark dataset exists for this task. To fill this gap, we build a specialized fursuit image dataset and present a three-stage hybrid pipeline Fursee for fursuit identity retrieval and clustering. First, YOLO detects and crops high-resolution fursuit head patches to improve localization of small and overlapping targets. Second, ArcFace optimizes DINOv3 embeddings to enlarge angular separation between different identities on the feature hypersphere. Third, DBSCAN performs unsupervised clustering, with silhouette-coefficient-driven search automatically selecting optimal hyperparameters rather than fixed manual radius. Retrieval and clustering experiments verify that our pipeline outperforms mainstream multimodal models including GPT5.5, Claude Opus 4.8 and Qwen3.7-Plus on all evaluation metrics, achieving competitive performance for fursuit head retrieval and grouping.
\end{abstract}    
\section{Introduction}
\label{sec:intro}

\par
The Furry Fandom represents a significant and growing subculture interested in anthropomorphic animal characters—animals with human characteristics such as bipedalism, speech, and facial expressions \cite{Gerbasi2008}. Central to this subculture is the \textbf{fursuit}, a custom-made costume that allows enthusiasts to physically embody their personas, or \textbf{fursonas}. Unlike standard cosplay, which replicates existing characters, fursuits are often unique, bespoke creations that serve as a primary medium for identity expression and social interaction within the community.
\par
In recent years, the surge in furry conventions worldwide has precipitated a sharp increase in fursuit-related imagery, as enthusiasts and photographers generate vast archives of convention photo sharing. However, the traditional workflow of manually sorting and tagging these photos is inefficient and labor-intensive, creating a significant bottleneck in distributing content to the respective owners. Consequently, there is an urgent demand for automated classification tools capable of intelligently routing these images to the correct furries without human intervention. Furthermore, the necessity for robust fursuit identity recognition cannot be ignored. There is a critical need for systems that can retrieve a specific user's identity from a database using a single fursuit photograph, thereby facilitating efficient identity verification and content indexing.
\par
Different from human faces and general objects, most fursuit heads share identical base molds, so their visual distinctions mainly lie in colors, patterns and complex fur textures rather than geometric structures or facial keypoints. Traditional face recognition methods relying on geometric features, such as Eigenfaces \cite{Turk1991Eigenfaces}, Fisherfaces \cite{belhumeur1996eigenfaces} and Local Binary Patterns (LBP) \cite{Ahonen2004LBP} for facial structure and texture modeling, fail in this scenario. Generic object detectors such as YOLO \cite{Redmon2016YOLO,Redmon2017YOLOv9000} can locate targets accurately, yet they lack discriminative power for fine-grained classification. Self-supervised models like DINOv3 \cite{Meta2025DINOv3} excel at extracting robust global and texture features, but direct whole-image classification is severely disturbed by cluttered backgrounds, occlusions and varying poses. Without domain-specific training on fursuit datasets, current visual systems achieve limited accuracy in fine-grained fursuit recognition. To address the above challenges, we propose a cascaded framework combining YOLO and DINOv3, and conduct domain-specific training on our newly built fursuit dataset. In the first stage, YOLO detects and crops fursuit head regions to remove background interference. In the second stage, DINOv3 encodes cropped fursuit patches into embeddings. In the third stage, we compute cosine similarity over embeddings to compare fursuit identities for downstream tasks. Our main contributions are summarized as follows:
\begin{itemize}
\item[$\bullet$] To the best of our knowledge, this work pioneers detection-guided embedding pipelines tailored for fine-grained fursuit identity retrieval and overlapping clustering. Existing YOLO-DINO hybrid methods are limited to object detection \cite{malaisree2025dino} and model distillation \cite{fourret2025improving}. Meanwhile, mainstream clustering and fine-grained recognition schemes merely adopt DINO \cite{maddigan2024re,markoff2025zero}, and no prior study explores such frameworks for fursuit-oriented fine-grained classification.
\item[$\bullet$] We design a hybrid YOLO+DINOv3 architecture, integrating the localization advantage of detectors and the strong representation capability of self-supervised Vision Transformers.
\item[$\bullet$] Extensive experiments demonstrate that our method outperforms mainstream models on the built dataset in terms of classification accuracy and robustness.
\end{itemize}
\section{Related work}
\label{sec:relawork}

\subsection{Fine-Grained Recognition and Metric Learning for Identity Discrimination}
\par
Fine-grained recognition targets subtle visual differences between visually homogeneous identities, which aligns with our fursuit discrimination scenario. Unlike human face recognition, a considerable number of fursuit heads share highly similar overall silhouettes, making unique color and texture patterns the core cues for distinguishing different identities. Early bilinear pooling methods capture local subtle cues for fine-grained classification \cite{Lin2015Bilinear}, yet they lack angular discriminative optimization for retrieval tasks.
\par
Metric learning is a core branch for identity retrieval, which constrains feature space to narrow intra-class distances and enlarge inter-class margins. ArcFace \cite{arcface} introduces angular margin penalties to boost discriminative embedding quality and has become the de facto standard for face and re-identification tasks. However, mainstream metric learning pipelines are designed for visible human faces or full-body pedestrians, and rarely address occlusion and multi-identity overlapping scenes as in fursuit imagery.
\par
Character recognition studies relying on clean official artwork datasets \cite{Kim2020AnimationChar} fail to generalize to real convention photos with heavy occlusion, dense overlapping and complex lighting, while dedicated visual benchmarks for fursuit identity retrieval remain unexplored.

\subsection{Person and Animal Re-Identification}
\par
Person re-identification (Re-ID) focuses on matching the same identity across multi-view images under varying poses and occlusions \cite{wang2020surveyreid}, sharing high similarity with our fursuit retrieval goal. Most existing Re-ID methods rely on visible human body or facial features, which are unavailable for fursuit heads. Animal Re-ID further extends fine-grained matching to furry textured subjects \cite{zheng2022animalreid}, but these datasets only contain single-animal frames without multi-identity overlapping, a critical challenge unique to furry convention photos.
\par
Self-supervised ViT backbones \cite{caron2021emerging} such as DINOv2/DINOv3 achieve strong texture-aware representation for Re-ID \cite{Oquab2023DINOv2,Meta2025DINOv3}, yet they suffer huge computational overhead when processing full high-resolution raw images crowded with small target heads. This motivates our detection-guided cropping strategy to extract critical local patches prior to feature embedding.

\subsection{Open-Set Unsupervised Identity Clustering}
\par
Real-world fursuit datasets contain unseen identities with unknown total categories, making supervised classification infeasible and calling for open-set clustering solutions. Advanced face clustering approaches such as STAR-FC \cite{shen2023star} and Ada-NETS \cite{Wang2022AdaNETS} have been widely applied to intelligent photo organization. Nevertheless, fursuit heads exhibit distinct visual traits from human faces, making existing face classification pipelines inapplicable. DBSCAN \cite{ester1996density} is widely adopted for density-based unsupervised grouping, and silhouette coefficients \cite{rousseeuw1987silhouette} serve as reliable unsupervised metrics to tune clustering hyperparameters without label supervision.

\subsection{Detection-Guided Embedding Hybrid Pipeline}
\par
The detect-then-recognize paradigm originates from two-stage detection frameworks \cite{Girshick2014RCNN,Ren2015FasterRCNN}, which decouple localization and feature extraction to avoid background interference. Modern lightweight one-stage detectors (YOLO series) deliver efficient target localization \cite{Redmon2017YOLOv9000}. We adopt YOLO solely to crop high-resolution fursuit head patches, addressing the small/dense target localization defects of raw full-image ViT inference.
\begin{figure*}[t!] 
    \centering
    \includegraphics[width=\textwidth]{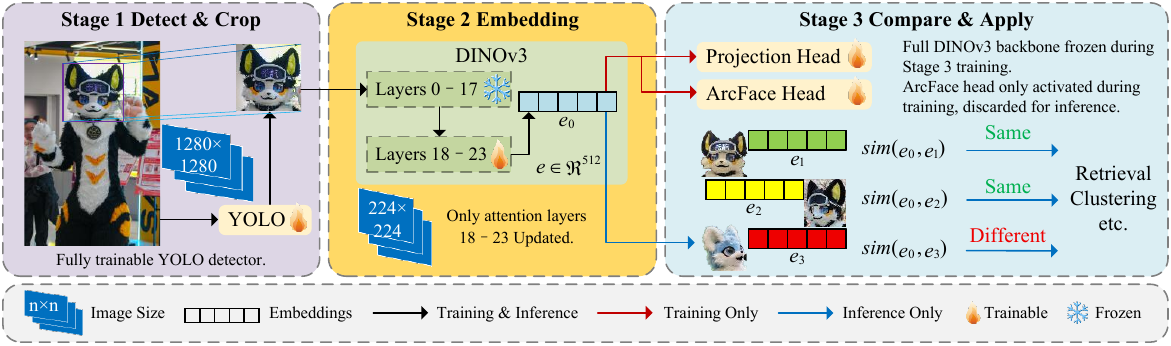} 
    \caption{The framework of our proposed method. Stage 1 trains YOLO to crop fursuit heads from high-resolution images. In Stage 2, DINOv3 layers 0–17 are frozen, only attention layers 18–23 are fine-tuned to extract embeddings. During Stage 3 training, DINOv3 backbone is fully frozen. Only projection and ArcFace heads are updated. ArcFace head is used solely for training and removed at inference, where embedding cosine similarity supports retrieval and clustering.}
    \label{fig:fursee_framework}
\end{figure*}

\section{Methodology}
\label{sec:Meth}

\subsection{Problem Formulation}
\par
We first define core sets from a set-theoretic perspective to formalize the fursuit recognition task. Let $\mathcal{P} = \{p_1, p_2, \dots, p_n\}$ be the universal set of all collected fursuit photos, and $\mathcal{F} = \{f_1, f_2, \dots, f_m\}$ denote the set of distinct fursuit identities appearing in $\mathcal{P}$. For each identity $f_j \in \mathcal{F}$, we define $S(f_j) = \{p \in \mathcal{P} \mid p \text{ contains fursuit identity } f_j\}$ as the subset of photos belonging to $f_j$. The fursuit retrieval task is formulated as
\[
\mathbf{Retrieval}(f_u) =
\begin{cases}
S(f_u) & \text{if } f_u \in \mathcal{F}, \\
\emptyset & \text{if } f_u \notin \mathcal{F},
\end{cases}
\]
which requires returning all photos containing the queried identity $f_u$ if $f_u$ exists in $\mathcal{F}$, and an empty set otherwise.
\par
In real-world scenarios, the total number of unique fursuit identities $m = |\mathcal{F}|$ is unknown in advance, making supervised classification impractical. We therefore formulate the fursuit identity clustering problem, which aims to assign each photo $p\in\mathcal{P}$ to all subsets corresponding to identities appearing within $p$, and collect all identity-related photo subsets 
\[
\mathbf{Clustering}(\mathcal{P}) = \{S(f_1), S(f_2), \dots, S(f_m)\}
\]
satisfying two constraints: completeness $\bigcup_{j=1}^m S(f_j) = \mathcal{P}$, and matching correctness that every photo $p \in S(f_j)$ must contain fursuit identity $f_j$. Different from exclusive single-label partitioning, subsets $S(f_i)$ and $S(f_j)$ for distinct identities $f_i\neq f_j$ allow non-empty intersections $S(f_i) \cap S(f_j) \neq \emptyset$, which naturally handles real-world photos capturing multiple fursuits simultaneously. The resulting collection of subsets serves as the unsupervised classification result, grouping all photos of each fursuit character without prior knowledge of the identity count.

\subsection{Overall Framework of our Method}
\par
We propose a three-stage cascaded pipeline for fursuit identity recognition. The core design of our framework converts the complicated multi-to-multi matching problem on multi-fursuit images into a series of simple single-instance comparison subtasks. After processing individual cropped patches to extract discriminative embeddings, we map all matching results back to the original full-resolution input. As illustrated in Figure \ref{fig:fursee_framework}, our pipeline applies separate parameter freezing rules across different training stages.
\par
In Stage 1, a fully trainable lightweight YOLO detector processes raw $1280\times1280$ photographs to locate all fursuit heads. Each detected head region is cropped into an independent patch containing exactly one fursuit character, which realizes the transformation from multi-target to single-instance matching. All cropped patches are uniformly resized to $224\times224$ for subsequent feature encoding.
\par
In Stage 2, we feed the standardized single-head patches into the pre-trained DINOv3 ViT backbone to produce compact 512-dimensional identity embeddings $e \in \mathbb{R}^{512}$. During Stage 2 fine-tuning, layers 0–17 of DINOv3 are completely frozen, while only attention layers 18–23 are optimized. This partial fine-tuning guides the model to concentrate its attention exclusively on fursuit head regions.
\par
In Stage 3, the entire DINOv3 backbone stays frozen during optimization. Only the projection head and ArcFace classification head are updated jointly. The ArcFace head is solely utilized to compute training losses and is removed in the inference phase. At test time, we calculate cosine similarity $\text{sim}(\cdot,\cdot)$ between embeddings to fulfill two practical downstream applications: fursuit image retrieval and unsupervised identity clustering.

\subsection{YOLO Detector Training}
\label{subsec:yolo_training}

\par
We adopt the YOLO26l \cite{jocher2026yolo26} detector as our stage-one localization backbone, with the input image resolution fixed to $1280 \times 1280$. This high-resolution setting mitigates detection degradation caused by tiny fursuit heads and densely arranged targets in crowded convention scenes.
\par
We construct a custom fursuit detection dataset by collecting real-scene photos from offline furry conventions, where each image is manually annotated with two types of labels: bounding boxes for fursuit head localization, and segmentation masks that are reused to supervise attention learning in the subsequent embedding stage (see Section \ref{subsec:attn-finetune}). To eliminate false positive detections where human facial regions are misclassified as fursuit heads, we introduce human face samples as hard negative training data, sourced from the public \textit{fairface-img-margin125-trainval} dataset \cite{Karkkainen2021fairface}.
\par
With fursuit and human face samples integrated, we partition the combined data into training and validation splits. The statistical scale of each subset is summarized in Table \ref{tab:detect-data-stat}. After full training, the detector gains the capability to distinguish two target categories: fursuit heads and human faces. During inference, we filter out all bounding boxes predicted for human faces, and only retain patches cropped from fursuit head regions as the input for the second-stage embedding module.

\begin{table}[h]
\centering
\caption{Data statistics of the YOLO training and validation sets, split by fursuit head and human face samples.}
\label{tab:detect-data-stat}
\begin{tabular}{lccc}
\hline
Category & Train Set & Validation Set & Total \\
\hline
Fursuit Head & 21800 & 2910 & 24710 \\
Human Face & 14105 & 1924 & 16029 \\
\hline
\end{tabular}
\end{table}

\subsection{Attention Module Fine-tuning}
\label{subsec:attn-finetune}
\par
YOLO-cropped image patches still contain redundant background regions. Therefore, it is necessary to fine-tune the attention module to suppress background interference and force the model to focus exclusively on fursuit heads. We perform attention fine-tuning using annotated fursuit head samples presented in the prior subsection (see Section \ref{subsec:yolo_training}). Each sample is annotated with both bounding boxes and segmentation masks, which serve as supervision signals for optimizing the attention mechanism of the pre-trained \textit{dinov3-vitl16-pretrain-lvd1689m} model \cite{Meta2025DINOv3}. All input images are resized to $224 \times 224$. With a patch size of $16 \times 16$, each image is divided into a $14 \times 14$ patch grid, and we construct a target attention heatmap $\mathcal{T} \in \mathbb{R}^{14 \times 14}$ to guide the fine-tuning process.
\par
First, we initialize a background matrix $\mathcal{T}_{bg}$ where all elements are set to the background score $\alpha_{bg} = 0.0$:
\begin{equation}
\mathcal{T}_{bg}[i,j] = 0.0, \quad \forall\, i,j \in \{1,2,\dots,14\}.
\end{equation}
\par
Second, we add scores for regions covered by bounding boxes. For patches within annotated bounding boxes, we superimpose a fixed score $\alpha_{bbox} = 0.2$:
\begin{equation}
\mathcal{T}_{bbox}[i,j] =
\begin{cases}
\mathcal{T}_{bg}[i,j] + 0.2, & (i,j) \in \mathcal{P}_B, \\
\mathcal{T}_{bg}[i,j], & \text{otherwise},
\end{cases}
\end{equation}
where $\mathcal{P}_B$ denotes the set of patches covered by bounding boxes. Third, we apply mask-based score assignment. Since segmentation masks provide more precise localization, patches covered by masks are overwritten with $\alpha_{mask} = 1.0$:
\begin{equation}
\mathcal{T}_{mask}[i,j] =
\begin{cases}
1.0, & (i,j) \in \mathcal{P}_M, \\
\mathcal{T}_{bbox}[i,j], & \text{otherwise},
\end{cases}
\end{equation}
where $\mathcal{P}_M$ represents patches covered by segmentation masks. Finally, we conduct L1 normalization to ensure the target distribution sums to one:
\begin{equation}
\mathcal{T}[i,j] = \frac{\mathcal{T}_{mask}[i,j]}{\sum_{x=1}^{14}\sum_{y=1}^{14} \mathcal{T}_{mask}[x,y]}.
\end{equation}
The overall target attention construction can be summarized as:
\begin{equation}
\mathcal{T} = \text{Normalize}\big(\mathcal{T}_{bg} + 0.2 \cdot \mathbb{I}_{\mathcal{P}_B} + 1.0 \cdot \mathbb{I}_{\mathcal{P}_M}\big),
\end{equation}
in which $\mathbb{I}(\cdot)$ is the indicator function.
\par
We adopt a cross-entropy loss derived from KL divergence to minimize the discrepancy between predicted attention distribution and the constructed target heatmap:
\begin{equation}
\mathcal{L}_{att} = -\sum_{i=1}^{14}\sum_{j=1}^{14} \mathcal{T}[i,j] \cdot \log\big(\mathcal{P}_{pred}[i,j] + \epsilon\big),
\end{equation}
where $\mathcal{P}_{pred}[i,j]$ is the Softmax-normalized predicted attention score, and $\epsilon$ is adopted to avoid numerical overflow. The attention scores are extracted from the specific layers of ViT, using the $\texttt{cls}$ token as the query for attention calculation: $\mathcal{P}_{pred} = \text{Softmax}\big(Q_{\texttt{cls}} \cdot K_{\text{patch}}^T\big)$. During optimization, gradient clipping is applied to prevent gradient explosion.

\begin{figure}[t]
    \centering
    \subfloat[The Original Image]{\includegraphics[width=0.15\textwidth]{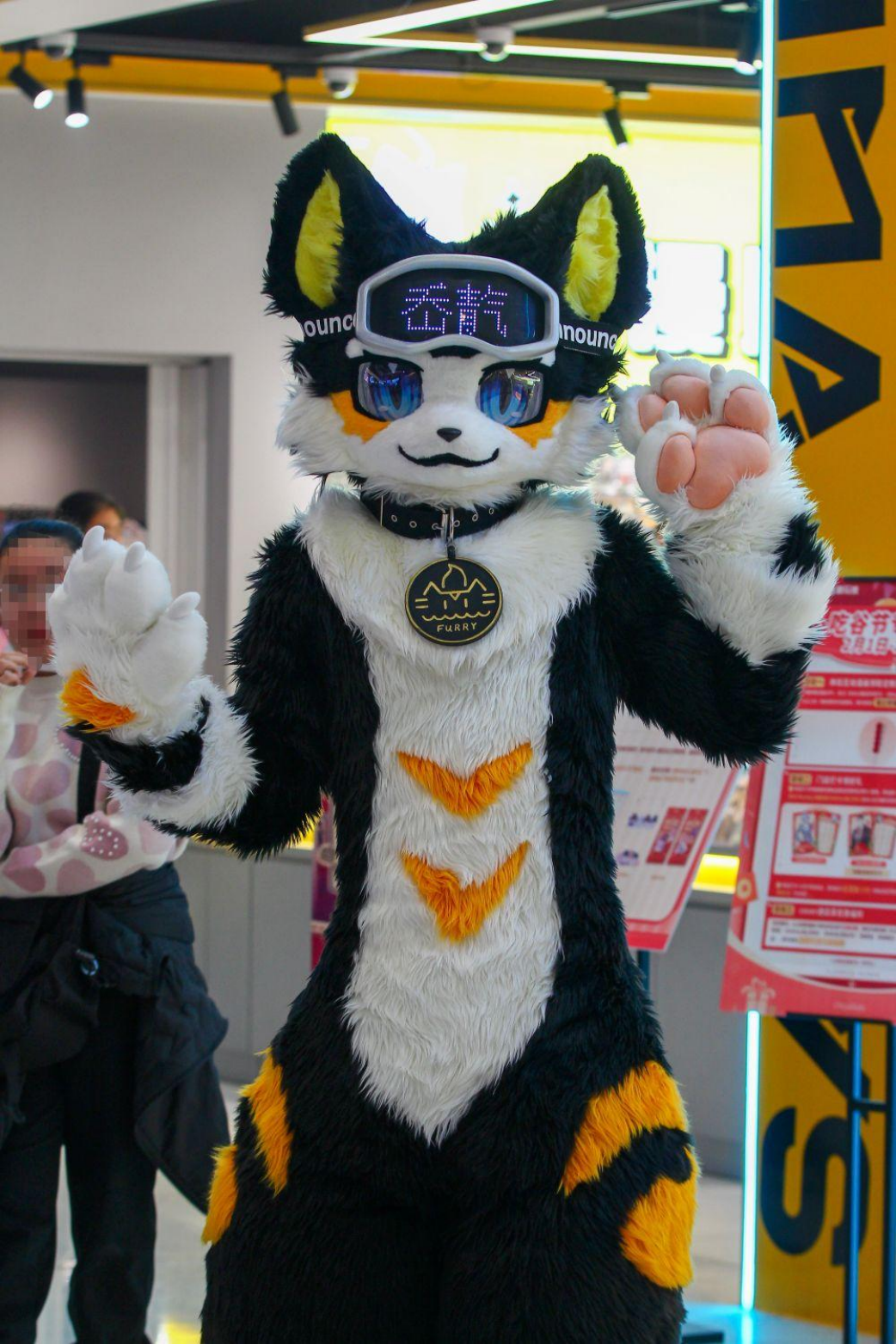}}
    \hfill
    \subfloat[Pre-fine-tuning]{\includegraphics[width=0.15\textwidth]{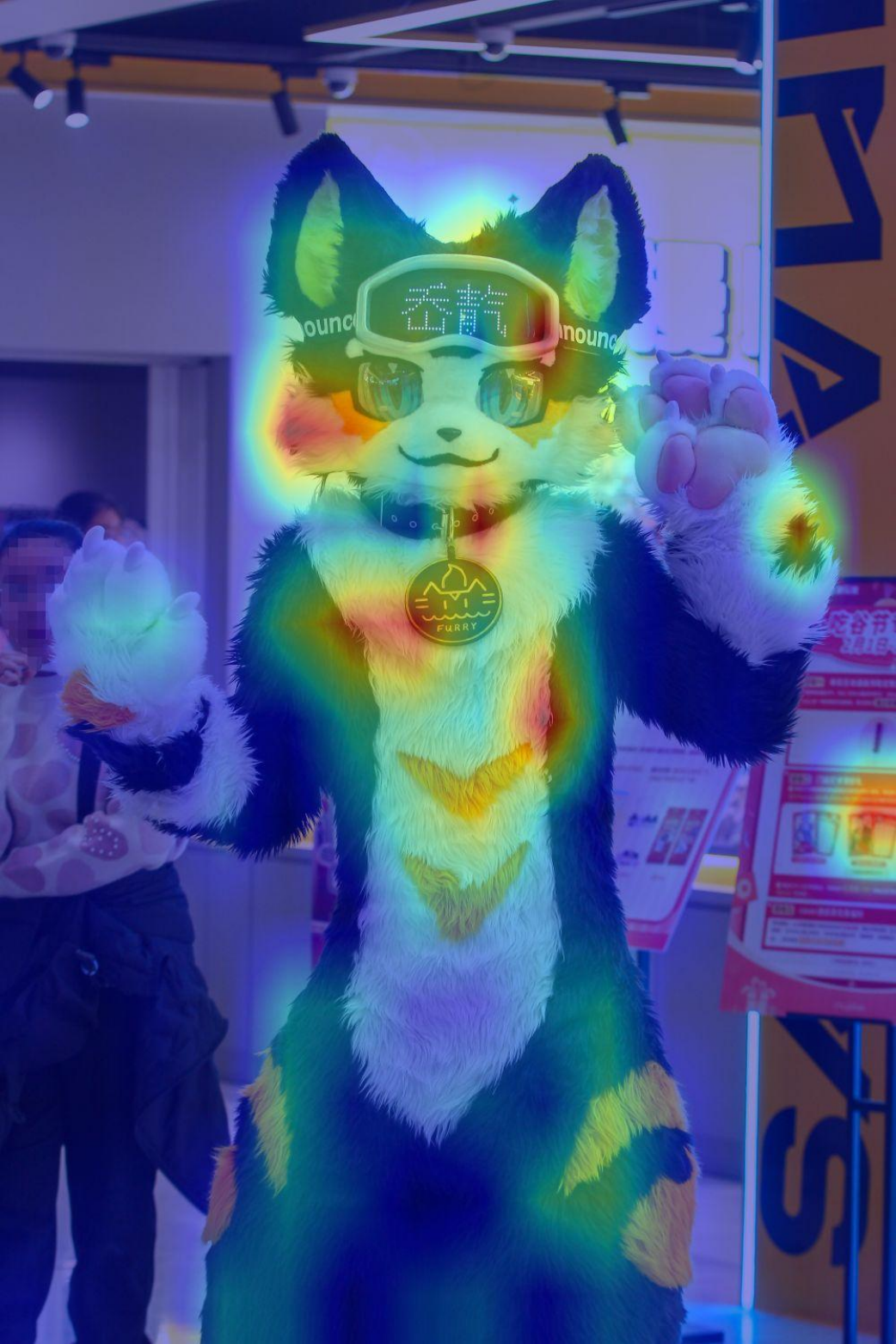}}
    \hfill
    \subfloat[Post-fine-tuning]{\includegraphics[width=0.15\textwidth]{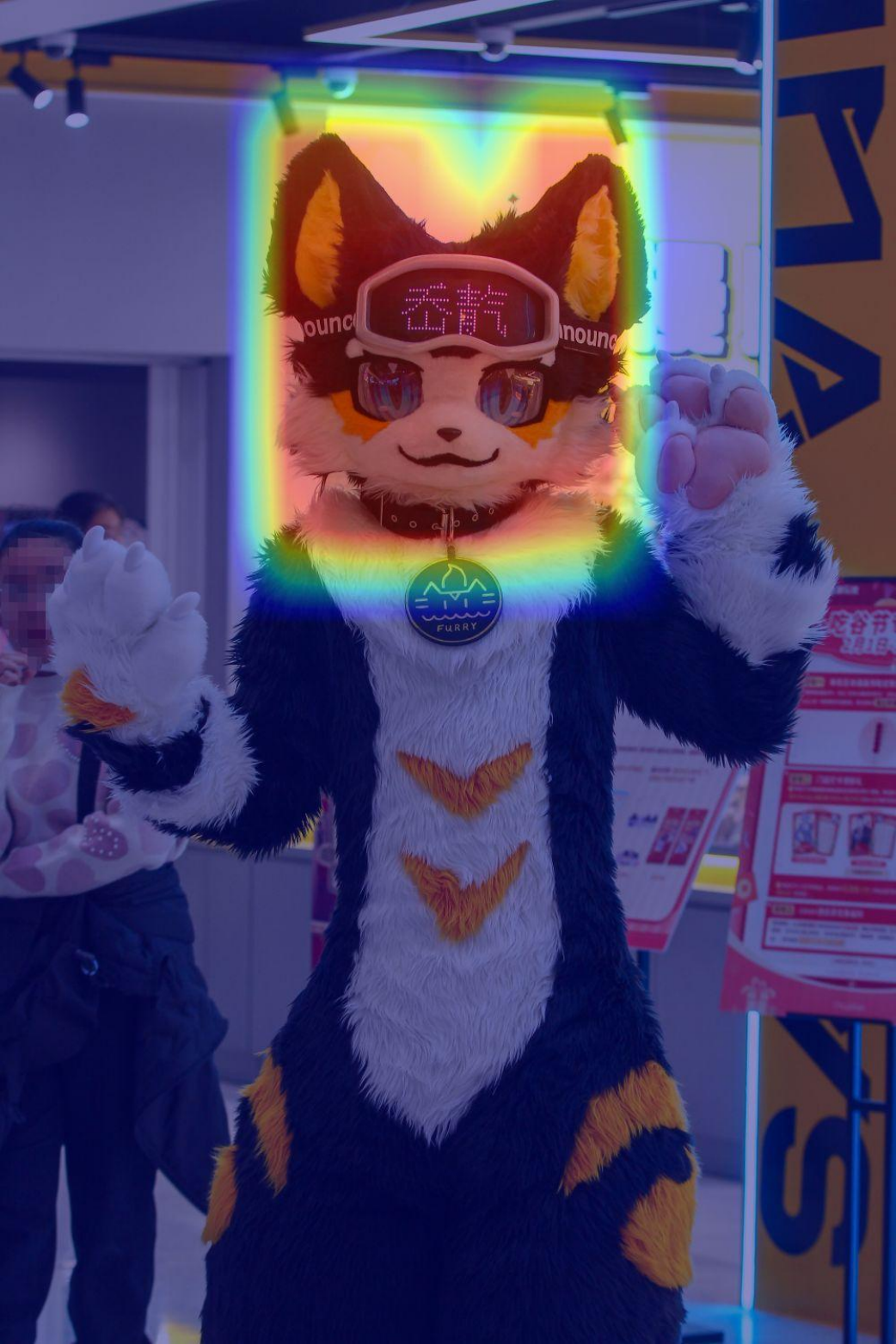}}
    \caption{Visualization of model attention heatmaps. Subfigure (a) shows the original input image for reference. (b) and (c) respectively display attention weights before and after fine-tuning. The proposed method enables the model to focus accurately on fursuit head areas.}
    \label{fig:attn_heatmap}
\end{figure}

\par
We visualize the original input image and attention distributions before and after fine-tuning in Figure \ref{fig:attn_heatmap}. Cool colors correspond to low attention weights, while warm colors represent high attention weights. Figure \ref{fig:attn_heatmap}(a) provides the raw fursuit photo as visual reference. Figure \ref{fig:attn_heatmap}(b) shows the attention map before fine-tuning, which presents scattered and unfocused responses across the whole body. After our targeted fine-tuning, the attention is tightly concentrated on the discriminative fursuit head region as shown in Figure \ref{fig:attn_heatmap}(c). The visualization results demonstrate that our supervision strategy effectively drives the model to focus on the key fursuit head features critical for identity discrimination.

\subsection{Embedding Learning with DINOv3}
\par
The raw fursuit images are first preprocessed by cropping individual fursuit heads from each picture, and all cropped head patches are manually labeled according to their corresponding identities. To alleviate overfitting and improve model generalization, we adopt a set of data augmentation strategies on the processed samples, including random rotation, horizontal and vertical mirror flipping, random shear transformation, random perturbations on hue, brightness and saturation, as well as radial and tangential lens distortion.
\par
We then train identity embeddings upon the DINOv3 backbone. During training, the entire DINOv3 backbone is frozen, and only the projection head and ArcFace classification head are optimized. We construct a hybrid loss composed of ArcFace loss \cite{arcface} and supervised contrastive loss \cite{khosla2020supervised}. We first expand the mathematical expression of each loss term separately.
\par
The ArcFace loss is defined as:
\begin{equation}
\mathcal{L}_{\text{ArcFace}} = -\frac{1}{N}\sum_{i=1}^N \log \frac{e^{s\cos(\theta_{y_i}+m)}}{e^{s\cos(\theta_{y_i}+m)} + \sum_{j \neq y_i} e^{s\cos\theta_{y_j}}}
\label{eq:arcface}
\end{equation}
where $N$ denotes the batch size, $s$ is the feature scaling factor, $m$ stands for the additive angular margin, $\theta_{y_i}$ is the angle between the normalized embedding of sample $i$ and the weight vector of its ground-truth identity, and $\theta_{y_j}$ refers to the angles between the sample embedding and all weight vectors of other negative identities. This loss enlarges angular margins between different identities in the feature space to strengthen embedding discrimination.
\par
The supervised contrastive loss is formulated as:
\begin{equation}
\mathcal{L}_{\text{SupCon}} = -\frac{1}{N}\sum_{i=1}^N \frac{1}{|P_i|}\sum_{p \in P_i} \log \frac{\exp(\boldsymbol{z}_i^\top \boldsymbol{z}_p / \tau)}{\sum_{k \neq i}\exp(\boldsymbol{z}_i^\top \boldsymbol{z}_k / \tau)}
\label{eq:supcon}
\end{equation}
where $\boldsymbol{z}$ represents normalized embedding vectors, $\tau$ is the temperature coefficient, and $P_i$ denotes the set of positive samples that share the same identity with anchor sample $i$. This loss pulls embeddings of identical identities closer and pushes cross-identity embeddings apart, bringing stronger generalization ability than standard Softmax loss.
\par
We combine the two loss terms with a balancing weight $\lambda$, and the final overall loss is written as:
\begin{equation}
\mathcal{L}_{\text{All}} = \mathcal{L}_{\text{ArcFace}} + \lambda \cdot \mathcal{L}_{\text{SupCon}}
\label{eq:total_loss}
\end{equation}
The weight coefficient $\lambda$ is set to 0.1 to balance the contribution of the two loss components.
\par
We utilize the PK Batch Sampler \cite{sohn2016improved} to construct training batches, where each batch contains $P$ distinct identities and $K$ samples per identity. This sampling scheme helps the model learn intra-class compactness and inter-class discrepancy, which is essential for metric learning \cite{bellet2013survey} based identity recognition tasks.
\par
\begin{figure}[htbp]
    \centering
    \includegraphics[width=0.45\textwidth]{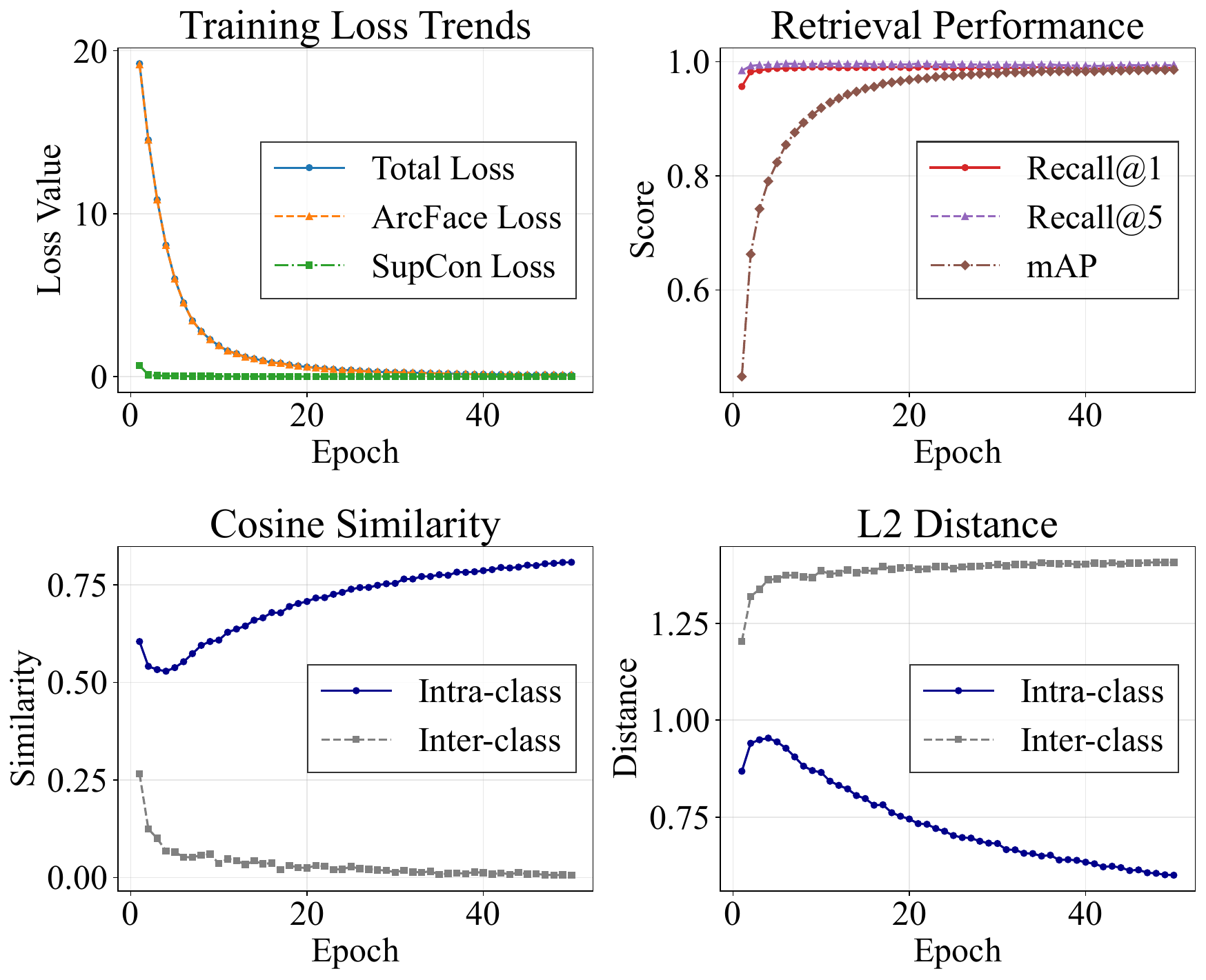}
    \caption{Training curves of the DINOv3-based embedding model.}
    \label{fig:training_performance}
\end{figure}

We evaluate the embedding quality on the validation set with comprehensive metrics, including training losses, retrieval accuracy, cosine similarity and L2 distance between feature vectors. All quantitative results across training epochs are presented in Figure \ref{fig:training_performance}.

\subsection{Identity Clustering with Adaptive DBSCAN}
\par
After extracting discriminative 512-dimensional embeddings from Stage 2, we perform identity clustering on all cropped fursuit head patches via DBSCAN \cite{ester1996density}, which serves as the core algorithm for the clustering branch in Stage 3 of our pipeline. The DINOv3 embedding model in Stage 2 is trained with ArcFace, which explicitly enlarges the angular margin between different identities on the hypersphere. Accordingly, we adopt cosine metric for clustering and silhouette evaluation. The silhouette coefficient calculated in the cosine space essentially measures whether the angular margins are well separated. With proper combinations of $\text{eps\_tolerance}$ and $\text{min\_samples}$, DBSCAN can partition samples along the high-density manifolds naturally formed by ArcFace features, and the silhouette coefficient will reach its maximum value at this optimal state. Instead of manually fixing a static radius $\epsilon$, we implement an automatic grid search strategy to select the optimal $\epsilon$ value within a predefined candidate interval, guided by the silhouette coefficient \cite{rousseeuw1987silhouette}.
\par
The silhouette coefficient $s(i)$ for a single embedding sample $i$ is formally defined as:
\begin{equation}
s(i) = \frac{b(i) - a(i)}{\max\big(a(i), b(i)\big)}
\label{eq:silhouette}
\end{equation}
where $a(i)$ denotes the average cosine distance between sample $i$ and all other samples within the same cluster, and $b(i)$ represents the minimum average cosine distance between sample $i$ and all samples from any other distinct cluster. The overall silhouette score of the whole dataset is the mean value of $s(i)$ over all samples, ranging from $-1$ to $1$. A higher silhouette score indicates tighter intra-cluster aggregation and clearer inter-cluster separation, which aligns perfectly with our fursuit identity discrimination target.
\par
We search for the optimal hyperparameter combination within a given range by maximizing the silhouette score. After assigning cluster labels to each cropped fursuit head patch, we map the clustering results back to the original full-resolution input images and categorize the original images into matched cluster folders.

\section{Experiments}
\label{sec:Exp}

\subsection{Dataset}
\par
Currently, to our knowledge, there is no publicly available standard dataset for fursuit-related visual tasks. Therefore, we construct a custom dataset for all experiments in this work. We collect 379 images captured from various furry conventions and manually divide them into 83 identity categories. Since multiple characters may appear in a single photograph, image overlap exists across different identity classes. This characteristic poses a challenge to the multi-object classification and clustering capability of the proposed pipeline.

\subsection{Evaluation Metrics}
We adopt different evaluation metrics for the retrieval task and clustering task respectively.

\subsubsection{Metrics for Retrieval Task}
For the fursuit image retrieval task, given a query image, the model retrieves the top-$k$ most similar samples from the whole dataset. We compare the retrieved results with the ground truth set which also contains $k$ matched samples, and calculate the hit rate as the core evaluation metric.

Let $\mathcal{R}$ denote the set of retrieved top-$k$ images, and $\mathcal{G}$ denote the ground truth set of matched images. The hit rate is defined as:
\begin{equation}
\text{HitRate} = \frac{|\mathcal{R} \cap \mathcal{G}|}{|\mathcal{G}|}
\end{equation}
where $|\cdot|$ represents the number of elements in a set. The numerator is the count of correctly matched samples, and the denominator is the total number of samples in the ground truth. A higher hit rate indicates better retrieval performance.

\subsubsection{Metrics for Clustering Task}
\par
Different from standard single-label clustering evaluation, our dataset contains images with multiple fursuit identities, which requires an overlapping-aware clustering evaluation pipeline. We adopt weighted precision, weighted recall and weighted F1-score with an additional structure coefficient to quantify clustering quality. The evaluation pipeline includes cluster matching based on IoU and the Hungarian algorithm \cite{kuhn1955hungarian}, class weight calculation, instance-level metric aggregation, and final structural correction.
\par
Let $\mathcal{G} = \{G_1, G_2, \dots, G_M\}$ denote the set of ground-truth identity clusters and $\mathcal{C} = \{C_1, C_2, \dots, C_N\}$ denote predicted clusters. We first compute the Intersection over Union (IoU) between each ground-truth cluster and predicted cluster:
\begin{equation}
\text{IoU}(G_i,C_j) =
\begin{cases}
\dfrac{|G_i \cap C_j|}{|G_i \cup C_j|}, & |G_i \cup C_j| > 0 \\
0, & \text{otherwise}
\end{cases}
\label{eq:iou_cluster}
\end{equation}
where $|\cdot|$ stands for the number of images contained in a set. We construct an IoU matrix and apply the Hungarian algorithm to minimize $1-\text{IoU}(G_i,C_j)$, achieving one-to-one optimal matching between ground-truth and predicted clusters, and building a mapping from each predicted cluster to its matched ground-truth identity.
\par
A structure coefficient term is introduced to punish inconsistency between the number of predicted clusters and ground-truth clusters:
\begin{equation}
\delta = \max\left(0,\; 1 - \left(\frac{|M-N|}{\max(M,N)}\right)^2\right)
\label{eq:structure_penalty}
\end{equation}
Here $M$ is the total number of ground-truth clusters, and $N$ is the number of predicted clusters. A larger gap between $M$ and $N$ produces a smaller structure coefficient, which suppresses the final clustering metrics.
\par
Each ground-truth identity $G_g$ is assigned a weight $w_g$ proportional to its total occurrence across all dataset images:
\begin{equation}
w_g = \frac{\text{Count}(G_g)}{\sum_{g'=1}^M \text{Count}(G_{g'})}
\label{eq:class_weight}
\end{equation}
$\text{Count}(G_g)$ records how many times identity $G_g$ appears in all annotated samples. Identities with more instances gain higher weights during global metric aggregation.
\par
For each ground-truth class $G_g$, we calculate class-wise precision $p_g$, recall $r_g$ and F1-score $f1_g$ based on the matched cluster mapping:
\begin{equation}
p_g = \frac{\text{TP}_g}{\text{TP}_g + \text{FP}_g}
\label{eq:cls_precision}
\end{equation}
\begin{equation}
r_g = \frac{\text{TP}_g}{\text{TP}_g + \text{FN}_g}
\label{eq:cls_recall}
\end{equation}
\begin{equation}
f1_g = \frac{2\cdot p_g \cdot r_g}{p_g + r_g}
\label{eq:cls_f1}
\end{equation}
where $\text{TP}_g$ denotes images correctly assigned to identity $G_g$, $\text{FP}_g$ denotes images falsely predicted as $G_g$, and $\text{FN}_g$ denotes images belonging to $G_g$ but not matched in predictions. For any ground-truth cluster that fails to match a corresponding predicted cluster, we assign $p_g=0$, $r_g=0$ and $f1_g=0$ to this category. Predicted clusters without matched ground-truth identities are excluded from metric aggregation, and only contribute to the calculation of the structure  coefficient term.
\par
We aggregate all class-level metrics weighted by identity weights to obtain raw global metrics:
\begin{align}
P_{\text{raw}} &= \sum_{g=1}^M w_g \cdot p_g \\
R_{\text{raw}} &= \sum_{g=1}^M w_g \cdot r_g \\
\text{F1}_{\text{raw}} &= \sum_{g=1}^M w_g \cdot f1_g
\end{align}
\par
Finally, the raw F1 metric is scaled by the structure coefficient $\delta$ to get the final evaluation indicator:
\begin{equation}
\text{F1}_{\text{final}} = \text{F1}_{\text{raw}} \cdot \delta
\label{eq:final_f1}
\end{equation}
The range of $\text{F1}_{\text{final}}$ is $[0,1]$. Values closer to 1 represent better overlapping clustering performance with consistent cluster quantity and accurate identity matching.

\subsection{Retrieval Experimental Results}
We compare our proposed Fursee model with three mainstream multimodal models, namely GPT5.5 \cite{openai2026gpt55sys}, Claude Opus 4.8 \cite{anthropic2026claude48sys} and Qwen3.7-Plus \cite{qwen2025qwen3}. All baseline models complete the fursuit retrieval task by receiving natural language task instructions and invoking their inherent visual understanding capabilities. We adopt hit rate as the evaluation metric. The quantitative results are presented in Table \ref{tab:retrieval_result}, where our Fursee model achieves state-of-the-art performance.

\begin{table}[htbp]
    \centering
    \caption{Retrieval performance comparison on fursuit retrieval task.}
    \label{tab:retrieval_result}
    \begin{tabular}{lc}
        \hline
        Model & Hit Rate \\
        \hline
        GPT5.5 & 70.00\% \\
        Claude Opus 4.8 & 85.00\% \\
        Qwen3.7-Plus & 46.67\% \\
        \hline
        \textbf {Fursee (Ours)} & \textbf{93.33\%} \\
        \hline
    \end{tabular}
\end{table}

\subsection{Clustering Experimental Results}
\par
We further conduct comparative clustering experiments using the same multimodal baseline models as the retrieval task. Instead of standard single-label clustering metrics, we adopt an overlapping-aware evaluation pipeline tailored for our multi-label fursuit dataset, which computes weighted raw precision $P_{\text{raw}}$, weighted raw recall $R_{\text{raw}}$, weighted raw F1-score $\text{F1}_{\text{raw}}$, and final corrected F1-score $\text{F1}_{\text{final}}$ after applying a structure penalty term. As illustrated in Table \ref{tab:clustering_result}, our proposed Fursee model obtains competitive performance and yields consistent improvements over all mainstream multimodal large model baselines on every evaluation metric, showing its promising effectiveness for fursuit head clustering.

\begin{table}[htbp]
    \centering
    \caption{Clustering performance comparison on fursuit clustering task.}
    \label{tab:clustering_result}
    \begin{tabular}{lcccc}
        \hline
        Model & $P_{\text{raw}}$ & $R_{\text{raw}}$ & $\text{F1}_{\text{raw}}$ &  $\text{F1}_{\text{final}}$ \\
        \hline
        GPT5.5 & 0.7601 & 0.5834 & 0.6077 & 0.6064 \\
        Claude Opus 4.8 & 0.7907 & 0.3983 & 0.4762 & 0.3956 \\
        Qwen3.7-Plus & 0.8779 & 0.6905 & 0.7442 & 0.7043 \\
        \hline
        \textbf{Fursee (Ours)} & \textbf{0.8986} & \textbf{0.8720} & \textbf{0.8760} & \textbf{0.8755} \\
        \hline
    \end{tabular}
\end{table}

\subsection{Ablation Study}
\par
To verify the necessity of the YOLO detection-and-crop stage in our pipeline, we conduct ablation experiments by removing the YOLO module and directly feeding full-resolution raw images into the DINOv3 ViT for retrieval and clustering tasks. Our DINOv3 embedding model is pre-trained to automatically focus semantic attention on fursuit head regions, which theoretically enables the model to complete identity retrieval and clustering without auxiliary object cropping.
\par
Table \ref{tab:ablation_yolo_retrieval} and Table \ref{tab:ablation_yolo_cluster} report the quantitative performance of the complete YOLO+DINO pipeline and the DINO-only variant.

\begin{table}[htbp]
    \centering
    \caption{Ablation results on retrieval task with/without YOLO module.}
    \label{tab:ablation_yolo_retrieval}
    \begin{tabular}{lc}
        \hline
        Pipeline & Hit Rate \\
        \hline
        DINO only & 86.67\% \\
        YOLO + DINO (Ours) & \textbf{93.33\%} \\
        \hline
    \end{tabular}
\end{table}

\begin{table}[htbp]
    \centering
    \caption{Ablation results on overlapping clustering task with/without YOLO module.}
    \label{tab:ablation_yolo_cluster}
    \begin{tabular}{lcc}
        \hline
        Metric & DINO only & YOLO + DINO (Ours) \\
        \hline
        $P_{\text{raw}}$ & \textbf{0.9452} & 0.8986 \\
        $R_{\text{raw}}$ & 0.8105 & \textbf{0.8720} \\
        $\text{F1}_{\text{raw}}$ & 0.8568 & \textbf{0.8760} \\
        $\delta$ & 0.9766 & \textbf{0.9994} \\
        $\text{F1}_{\text{final}}$ & 0.8368 & \textbf{0.8755} \\
        \hline
    \end{tabular}
\end{table}
\par
From the results, we observe that the DINO-only variant achieves slightly higher weighted raw precision but suffers clear degradation on recall and final corrected F1-score. The core limitation lies in the input resolution of DINOv3 ViT. Scaling the ViT input resolution to match the cropped high-resolution patches from YOLO introduces prohibitive computational overhead. Without YOLO cropping, the model exhibits weak localization capability for small and densely packed fursuit head targets, which degrades overall retrieval and clustering performance. This ablation validates the indispensable role of the YOLO detection stage in our pipeline.
\section{Conclusion}
\par
We propose a three-stage pipeline dubbed Fursee: the YOLO detection module extracts high-resolution fursuit head crops to mitigate poor localization on small and dense targets, DINOv3 embeddings optimized with ArcFace generate discriminative angular features, and DBSCAN clustering with silhouette coefficient guided automatic hyperparameter search partitions identity manifolds without manually fixed radius. To fairly evaluate multi-label overlapping clustering results inconsistent with standard single-label metrics, we design a dedicated evaluation pipeline combining IoU Hungarian matching and a cluster quantity structure penalty to compute weighted precision, recall and corrected final F1-score.
\par
Comparative experiments on retrieval and clustering tasks demonstrate that our Fursee pipeline achieves improvements over mainstream multimodal large models including GPT5.5, Claude Opus 4.8 and Qwen3.7-Plus across all evaluation metrics. Ablation studies further validate the necessity of the YOLO cropping stage; the DINO-only variant suffers degraded recall and final clustering performance due to limited native ViT resolution and insufficient capability for crowded small targets, despite marginally higher raw precision.
\par
This work still has several limitations. Our dataset only covers limited regional fursuit styles, and the model performance declines severely under heavy occlusion or low-quality blurry inputs. Additionally, the current model tends to split the same fursuit identity into separate clusters when the character wears distinct accessories, which impairs clustering consistency. In future work, we plan to expand the dataset with diverse cross-regional samples and extreme hard cases, and mitigate the accessory-sensitive clustering bias in subsequent research.
{
    \small
    \bibliographystyle{ieeenat_fullname}
    \bibliography{main}

@String(PAMI = {IEEE Trans. Pattern Anal. Mach. Intell.})

@String(CVPR= {IEEE Conf. Comput. Vis. Pattern Recog.})

@String(ICCV= {Int. Conf. Comput. Vis.})

@String(ECCV= {Eur. Conf. Comput. Vis.})

@String(NIPS= {Adv. Neural Inform. Process. Syst.})

@String(TMM  = {IEEE Trans. Multimedia})

@String(ICIP = {IEEE Int. Conf. Image Process.})

@String(ICLR = {Int. Conf. Learn. Represent.})

@String(PAMI  = {IEEE TPAMI})

@String(CVPR  = {CVPR})

@String(ICCV  = {ICCV})

@String(ECCV  = {ECCV})

@String(NIPS  = {NeurIPS})

@String(TMM   =	{IEEE TMM})

@String(ICIP  = {ICIP})

@String(ICLR  = {ICLR})

@article{Gerbasi2008,
  author = {Bernstein, Penny and Paolone, Nicholas and Higner, Justin and Gerbasi, Kathleen and Conway, Samuel and Privitera, Adam and Scaletta, Laura},
  title = {Furries From A to Z (Anthropomorphism to Zoomorphism)},
  journal = {Society \& Animals},
  pages = {197--222},
  year = {2008}
}

@inproceedings{Redmon2016YOLO,
  author = {Redmon, Joseph and Divvala, Santosh and Girshick, Ross and Farhadi, Ali},
  title = {You Only Look Once: Unified, Real-Time Object Detection},
  booktitle = CVPR,
  pages = {779--788},
  year = {2016}
}

@inproceedings{Redmon2017YOLOv9000,
  author = {Redmon, Joseph and Farhadi, Ali},
  title = {YOLO9000: Better, Faster, Stronger},
  booktitle = CVPR,
  pages = {7263--7271},
  year = {2017}
}

@inproceedings{caron2021emerging,
  title={Emerging Properties in Self-Supervised Vision Transformers},
  author={Caron, Mathilde and Touvron, Hugo and Misra, Ishan and J{\'e}gou, Herv{\'e} and Mairal, Julien and Bojanowski, Piotr and Joulin, Armand},
  booktitle=ICCV,
  pages={9650--9660},
  year={2021}
}

@article{oquab2023dinov2,
  title={Dinov2: Learning robust visual features without supervision},
  author={Oquab, Maxime and Darcet, Timoth{\'e}e and Moutakanni, Th{\'e}o and Vo, Huy and Szafraniec, Marc and Khalidov, Vasil and Fernandez, Pierre and Haziza, Daniel and Massa, Francisco and El-Nouby, Alaaeldin and others},
  journal={arXiv preprint arXiv:2304.07193},
  year={2023}
}

@article{Meta2025DINOv3,
  title={Dinov3},
  author={Sim{\'e}oni, Oriane and Vo, Huy V and Seitzer, Maximilian and Baldassarre, Federico and Oquab, Maxime and Jose, Cijo and Khalidov, Vasil and Szafraniec, Marc and Yi, Seungeun and Ramamonjisoa, Micha{\"e}l and others},
  journal={arXiv preprint arXiv:2508.10104},
  year={2025}
}

@inproceedings{Lin2015Bilinear,
  author = {Lin, Tsung-Yu and RoyChowdhury, Aruni and Maji, Subhransu},
  title = {Bilinear CNN Models for Fine-Grained Visual Recognition},
  booktitle = ICCV,
  pages = {1449--1457},
  year = {2015}
}

@inproceedings{Girshick2014RCNN,
  author = {Girshick, Ross and Donahue, Jeff and Darrell, Trevor and Malik, Jitendra},
  title = {Rich Feature Hierarchies for Accurate Object Detection and Semantic Segmentation},
  booktitle = CVPR,
  pages = {580--587},
  year = {2014}
}

@inproceedings{Ren2015FasterRCNN,
  author = {Ren, Shaoqing and He, Kaiming and Girshick, Ross and Sun, Jian},
  title = {Faster R-CNN: Towards Real-Time Object Detection with Region Proposal Networks},
  booktitle = NIPS,
  volume={28},
  year = {2015}
}

@inproceedings{Turk1991Eigenfaces,
  author = {Turk, Matthew A and Pentland, Alex and others},
  title = {Face Recognition Using Eigenfaces},
  booktitle = CVPR,
  pages = {586--591},
  year = {1991}
}

@inproceedings{belhumeur1996eigenfaces,
  title={Eigenfaces vs. Fisherfaces: Recognition Using Class Specific Linear Projection},
  author={Belhumeur, Peter N and Hespanha, Joao P and Kriegman, David J},
  booktitle=ECCV,
  pages={43--58},
  year={1996},
  organization={Springer}
}

@inproceedings{Ahonen2004LBP,
  author = {Ahonen, Timo and Hadid, Abdenour and Pietikainen, Matti},
  title = {Face Recognition with Local Binary Patterns},
  booktitle = ECCV,
  pages = {469--481},
  year = {2004}
}

@article{malaisree2025dino,
  title={DINO-YOLO: Self-Supervised Pre-training for Data-Efficient Object Detection in Civil Engineering Applications.},
  author={Malaisree, P and Youwai, Sompote and Kitkobsin, Tipok and Janrungautai, S and Amorndechaphon, D and Rojanavasu, P},
  journal={arXiv preprint arXiv: 2510.25140},
  year={2025}
}

@inproceedings{fourret2025improving,
  title={Improving Yolov8 For Fast Few-Shot Object Detection By Dinov2 Distillation},
  author={Fourret, Guillaume and Chaumont, Marc and Fiorio, Christophe and Subsol, G{\'e}rard},
  booktitle=ICIP,
  pages={803--808},
  year={2025},
  organization={IEEE}
}

@inproceedings{maddigan2024re,
  title={Re-Identification of Individual Kākā: An Explainable DINO-Based Mode},
  author={Maddigan, Paula and Ehrhardt, Oskar and Lensen, Andrew and Shaw, Rachael C},
  booktitle={2024 39th International Conference on Image and Vision Computing New Zealand (IVCNZ)},
  pages={1--6},
  year={2024},
  organization={IEEE}
}

@article{markoff2025zero,
  title={Zero-Shot Wildlife Sorting Using Vision Transformers: Evaluating Clustering and Continuous Similarity Ordering},
  author={Markoff, Hugo and Galaktionovs, Jevgenijs},
  journal={arXiv preprint arXiv:2510.14596},
  year={2025}
}

@article{shen2023star,
  title={STAR-FC: Structure-Aware Face Clustering on Ultra-Large-Scale Graphs},
  author={Shen, Shuai and Li, Wanhua and Zhu, Zheng and Zhou, Jie and Lu, Jiwen},
  journal=PAMI,
  volume={45},
  number={11},
  pages={14005--14019},
  year={2023},
  publisher={IEEE}
}

@inproceedings{Wang2022AdaNETS,
  author={Wang, Yaohua and Zhang, Yaobin and Zhang, Fangyi and Wang, Senzhang and Lin, Ming and Zhang, YuQi and Sun, Xiuyu},
  title = {Ada-NETS: Face Clustering via Adaptive Neighbour Discovery in the Structure Space},
  booktitle = ICLR,
  year = {2022}
}

@article{Kim2020AnimationChar,
  author={Kim, Hayeon and Lee, Eun-Cheol and Seo, Yongseok and Im, Dong-Hyuck and Lee, In-Kwon},
  title = {Character Detection in Animated Movies Using Multi-Style Adaptation and Visual Attention},
  journal = TMM,
  volume = {23},
  pages = {1990--2004},
  year = {2020}
}

@inproceedings{Karkkainen2021fairface,
  author = {Karkkainen, Kimmo and Joo, Jungseock},
  title = {FairFace: Face Attribute Dataset for Balanced Race, Gender, and Age for Bias Measurement and Mitigation},
  booktitle={Proceedings of the IEEE/CVF winter conference on applications of computer vision},
  pages={1548--1558},
  year={2021}
}

@misc{jocher2026yolo26,
  title={Ultralytics YOLO26: Unified Real-Time End-to-End Vision Models},
  author={Jocher, Glenn and Qiu, Jing and Liu, Mengyu and Lyu, Shuai and Akyon, Fatih Cagatay and Kalfaoglu, Muhammet Esat},
  year={2026},
  eprint={2606.03748},
  archivePrefix={arXiv},
  primaryClass={cs.CV},
  doi={10.48550/arXiv.2606.03748},
  url={https://arxiv.org/abs/2606.03748}
}

@inproceedings{arcface,
 title={ArcFace: Additive Angular Margin Loss for Deep Face Recognition},
 author={Deng, Jiankang and Guo, Jia and Xue, Niannan and Zafeiriou, Stefanos},
 booktitle=CVPR,
 year={2019},
 pages={4690--4699}
}

@inproceedings{khosla2020supervised,
  title={Supervised Contrastive Learning},
  author={Khosla, Prannay and Teterwak, Piotr and Wang, Chen and Sarna, Aaron and Tian, Yonglong and Isola, Phillip and Maschinot, Aaron and Liu, Ce and Krishnan, Dilip},
  booktitle=NIPS,
  volume={33},
  pages={18661--18673},
  year={2020}
}

@article{bellet2013survey,
  title={A Survey on Metric Learning for Feature Vectors and Structured Data},
  author={Bellet, Aurelien and Habrard, Amaury and Sebban, Marc},
  journal={arXiv preprint arXiv:1306.6709},
  year={2013}
}

@article{sohn2016improved,
  title={Improved Deep Metric Learning with Multi-class N-pair Loss Objective},
  author={Sohn, Kihyuk},
  journal={Advances in neural information processing systems},
  volume={29},
  year={2016}
}

@inproceedings{ester1996density,
  title={A density-based algorithm for discovering clusters in large spatial databases with noise},
  author={Ester, Martin and Kriegel, Hans-Peter and Sander, J{\"o}rg and Xu, Xiaowei and others},
  booktitle={KDD},
  volume={96},
  number={34},
  pages={226--231},
  year={1996}
}

@article{rousseeuw1987silhouette,
  title={Silhouettes: a graphical aid to the interpretation and validation of cluster analysis},
  author={Rousseeuw, Peter J},
  journal={Journal of computational and applied mathematics},
  volume={20},
  pages={53--65},
  year={1987},
  publisher={Elsevier}
}

@techreport{openai2026gpt55sys,
  title={GPT-5.5 System Card},
  author={OpenAI},
  institution={OpenAI},
  year={2026},
  month={4},
  url={https://deploymentsafety.openai.com/gpt-5-5/gpt-5-5.pdf}
}

@techreport{anthropic2026claude48sys,
  title={System Card: Claude Opus 4.8},
  author={Anthropic},
  institution={Anthropic},
  year={2026},
  month={5},
  url={https://cdn.sanity.io/files/4zrzovbb/website/c886650a2e96fc0925c805a1a7ca77314ccbf4a6.pdf}
}

@article{qwen2025qwen3,
  title={Qwen3 Technical Report},
  author={Qwen Team},
  journal={arXiv preprint arXiv:2505.09388},
  year={2025}
}

@article{kuhn1955hungarian,
  title={The Hungarian method for the assignment problem},
  author={Kuhn, Harold W},
  journal={Naval research logistics quarterly},
  volume={2},
  number={1-2},
  pages={83--97},
  year={1955},
  publisher={Wiley Online Library}
}

@article{wang2020surveyreid,
  title={A Comprehensive Overview of Person Re-Identification Approaches},
  author={Wang, Hongbo and Du, Haomin and Zhao, Yue and Yan, Jiming},
  journal={IEEE Access},
  volume={8},
  pages={45556--45583},
  year={2020},
  publisher={IEEE}
}

@article{zheng2022animalreid,
  title={Wild Terrestrial Animal Re-Identification Based on an Improved Locally Aware Transformer with a Cross-Attention Mechanism},
  author={Zheng, Zhaoxiang and Zhao, Yaqin and Li, Ao and Yu, Qiuping},
  journal={Animals},
  volume={12},
  number={24},
  pages={3503},
  year={2022},
  publisher={MDPI}
}
}


\end{document}